\documentclass{Interspeech}

\interspeechcameraready

\title{Large Language Models based ASR Error Correction for Child Conversations}

\author[affiliation={1}]{Anfeng}{Xu$^*$}
\author[affiliation={1}]{Tiantian}{Feng$^*$}
\author[affiliation={2}]{So Hyun}{Kim}
\author[affiliation={3}]{Somer}{Bishop}
\author[affiliation={4}]{Catherine}{Lord}
\author[affiliation={1}]{Shrikanth}{Narayanan}

\affiliation{Viterbi School of Engineering}{University of Southern California}{USA}
\affiliation{School of Psychology}{Korea University}{South Korea}
\affiliation{Weill Institute for Neurosciences}{University of California, San Francisco}{USA}
\affiliation{Semel Institute of Neuroscience and Human Behavior}{University of California, Los Angeles}{USA}
\email{anfengxu@usc.edu}

\usepackage{comment}
\usepackage{multirow}
\begin{document}

\maketitle
\def\thefootnote{*}\footnotetext{These authors contributed equally to this work}\def\thefootnote{\arabic{footnote}}
\begin{abstract}
    Automatic Speech Recognition (ASR) has recently shown remarkable progress, but accurately transcribing children's speech remains a significant challenge. Recent developments in Large Language Models (LLMs) have shown promise in improving ASR transcriptions. However, their applications in child speech including conversational scenarios are under-explored. In this study, we explore the use of LLMs in correcting ASR errors for conversational child speech. We demonstrate the promises and challenges of LLMs through experiments on two children's conversational speech datasets with both zero-shot and fine-tuned ASR outputs. We find that while LLMs are helpful in correcting zero-shot ASR outputs and fine-tuned CTC-based ASR outputs, it remains challenging for LLMs to improve ASR performance when incorporating contextual information or when using fine-tuned autoregressive ASR (e.g., Whisper) outputs.
\end{abstract}

\section{Introduction}
Automatic Speech Recognition (ASR) has made substantial advances in recent years, driven by Speech Foundation Models (SFM) \cite{bommasani2021opportunities}, trained with advanced deep learning architectures, such as transformers \cite{vaswani2017u} and conformers \cite{gulati20_interspeech}, while leveraging extensive training data.
SFMs can be categorized into two categories. The first is end-to-end supervised models that leverage massive labeled datasets to jointly align acoustic and language information. Whisper \cite{radford2023robust} and Parakeet \cite{rekesh2023fast} are widely used models in this category.
The second is models trained with self-supervised learning (SSL) regime, such as Wav2vec 2.0 \cite{baevski2020wav2vec}, HuBERT \cite{hsu2021hubert}, and WavLM \cite{chen2022wavlm}, which learn speech representations from unlabeled audio data. When fine-tuned, the SSL-based models demonstrate competitive performance on ASR tasks.

However, technological improvements have primarily focused on improving adult speech recognition, while ASR for children's speech remains a persistent challenge: recent evaluations have shown ASR error rates for child speech are 10 to 19 times worse than for adults with general models, and 6 times worse despite adaptation on children speech \cite{GURUNATHSHIVAKUMAR2022101289}. Accurately transcribing conversations involving children is crucial for various applications, including educational technology, clinical assessments, and developmental research \cite{fan24b_interspeech}. Yet, current SFMs underperform on this task, as children's speech patterns differ significantly from those of adults in terms of acoustic-phonetic characteristics \cite{Lee1999Acousticsofchildrensspeech:,lee2014developmental}, vocabulary usage, prosodic features, and conversational dynamics \cite{1255448}. These challenges are further compounded by the relative scarcity of large-scale and naturalistic children's speech datasets, resulting in ASR systems that fall short of generalizing to child-inclusive applications \cite{10447428,GURUNATHSHIVAKUMAR2022101289}.

\begin{figure}[!t]
  \centering
  \centerline{\includegraphics[width=0.9\linewidth]{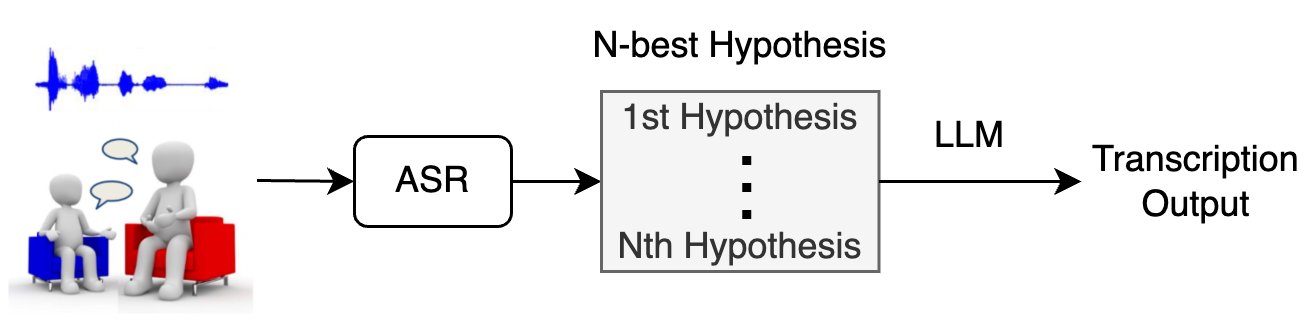}}  
  \caption{Overall pipeline for ASR with LLM error correction.}\medskip
  \label{fig:pipeline}
  \vspace{-5mm}
\end{figure}

Large Language Models (LLMs) have gained substantial attention in natural language processing through their advanced capabilities in processing large volumes of input data and generating sophisticated inferences and responses \cite{zhao2023survey}. While initially developed for text-related tasks, these models have shown promising applications in ASR systems. For example, several studies~\cite{fathullah2024prompting, ma2024embarrassingly, 10445874} have shown that LLMs can perform ASR tasks when integrated with audio or speech encoders. Ogawa et al.~\cite{ogawa2024applying} used LLMs to improve ASR transcriptions by rescoring ASR hypotheses using conversational contexts.

LLMs' capability to process language structure, context, and semantic relationships enables them to effectively correct ASR errors by considering both narrow linguistic patterns and broader semantic context \cite{zhao2023survey}. Recent works \cite{chen2024hyporadise, ma2023can} have demonstrated the effectiveness of leveraging LLMs for ASR error correction by selecting and refining the most probable transcription using N-best hypotheses from ASR systems. However, to our knowledge, limited work examines LLMs for ASR error correction in conversational settings, particularly in child-inclusive contexts. Additionally, these works have mainly focused on improving zero-shot ASR outputs, while LLMs' capabilities to improve fine-tuned ASR outputs remain unclear.

In this work, we investigate approaches for improving ASR accuracy with child-adult conversations by error correction using LLMs. Our method builds upon the HyPoradise \cite{chen2024hyporadise} benchmark for LLM-based ASR error correction by adapting it to handle conversational child speech. The main contributions of this work are summarized below.

\begin{itemize}
    \item We incorporate LLMs for ASR error correction for children's conversational speech. To the best our our knowledge, this work is one of the earliest attempts in this domain.
    \item We investigate the effectiveness of LLMs in improving ASR transcription across multiple scenarios: applying LLMs to both zero-shot and fine-tuned outputs, from supervised and self-supervised ASR models.
    \item We investigate utilizing conversational context to improve LLM-based error correction by leveraging previous utterances in the conversation history as additional input.
\end{itemize}

\section{Methods}
\subsection{LLM Prompt Design}
Our approach to LLM-based error correction for child-adult conversations builds upon the benchmark framework established by HyPoradise~\cite{chen2024hyporadise}, which uses N-best hypotheses from ASR for LLMs as illustrated in Figure~\ref{fig:pipeline}. While HyPoradise focused on general ASR error correction, we specifically adapted their methodology for conversational speech between children and adults. We use 5-best hypotheses from ASR outcomes and train an LLM for error correction. We use LLaMA3 \cite{dubey2024llama} models for our experiments.

\subsubsection{ASR Error Correction without Context}
\begin{figure}[]
  \centering
  \centerline{\includegraphics[width=0.8\linewidth]{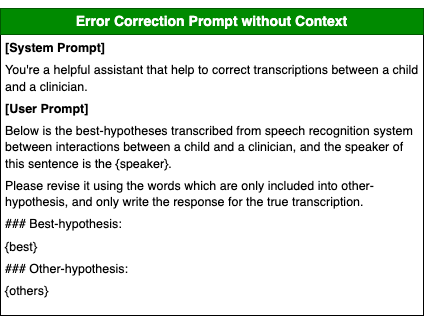}}
  \vspace{-3mm}
  \caption{LLM prompt without context.}\medskip
  \label{fig:no_context}
  \vspace{-5mm}
\end{figure}

We first examine if LLMs can help correct ASR errors without previous conversational context. The prompt we use is shown in Figure~\ref{fig:no_context}. \textit{\{speaker\}} is replaced with either ``Child" or ``Adult", while \textit{\{best\}} and \textit{\{others\}} are replaced with the top-1 ASR hypothesis and remaining top ASR hypotheses, respectively. 

\subsubsection{ASR Error Correction with Context}
\begin{figure}[]
  \centering
  \centerline{\includegraphics[width=0.8\linewidth]{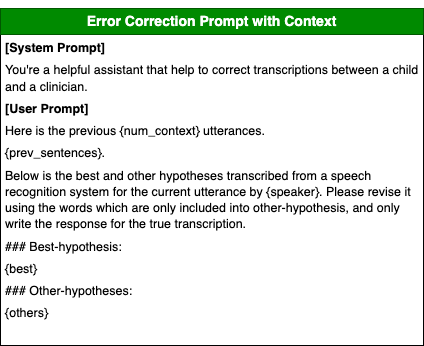}}
  \vspace{-3mm}
  \caption{LLM prompt with context.}\medskip
  \label{fig:context}
  \vspace{-5mm}
\end{figure}

Our previous work has shown the utility of dialog context modeling to improve child ASR \cite{Kumar2020LeveragingLinguisticContextin}. We hence investigate whether incorporating prior conversational context can also guide LLMs to correct ASR errors in child-adult interactions. We also note that young children frequently echo adults' speech in conversational interaction, providing natural repetition that could serve as meaningful contextual information for error detection. We experiment by including either 1 or 3 previous utterances. Figure~\ref{fig:context} shows the prompt we used. \textit{\{speaker\}}, \textit{\{best\}}, and \textit{\{others\}} are replaced similarly as for the case without context. \textit{\{num\_context\}} is replaced with the number of previous utterances, and \textit{\{prev\_sentences\}} is replaced by the previous 1 or 3 utterances with speaker tags (e.g., ``Adult: how are you?"). The ground-truth previous utterances are used during the training, while inferred previous utterances are used for testing. When there is no previous utterance, it is replaced by ``None."

\subsection{ASR Models}
To evaluate the effectiveness of LLM-based ASR error correction, we conduct experiments using two distinct ASR architectures to assess our approach across different ASR paradigms. 

\textbf{Whisper} \cite{radford2023robust} uses an attention-based encoder-decoder architecture, trained jointly for ASR and related tasks (e.g., translation) with weak supervision, using 680k hours of multi-language speech content gathered from online sources. It has shown competitive results on ASR benchmarks and works robustly across different recording scenarios. Specifically, we apply the Whisper-small (\textbf{WSP-S}), Whisper-large-v3 (\textbf{WSP-L}), and Whisper-large-v3-turbo (\textbf{WSP-L-T}) models. We use all these models to generate zero-shot ASR outputs and report the fine-tuned ASR results with the WSP-L-T. During inference, we apply beam search decoding with a beam size of 60.

\textbf{WavLM} \cite{chen2022wavlm} is an SSL-based model that builds upon Wav2vec 2.0 and HuBERT. It learns universal speech representations through masked speech prediction in pre-training. It shows strong transfer learning performance across diverse speech processing tasks, thus being the current state-of-the-art for the SUPERB benchmark \cite{yang2021superb}. We only use fine-tuned WavLM outputs for ASR experiments rather than zero-shot inference. In our experiments, we fine-tune WavLM-large (\textbf{WavLM-L}) with CTC loss for character predictions. For WavLM inference, we employ beam search decoding with a smaller beam size of 10.

\subsection{ASR Fine-tuning}
In addition to LLM-based error correction for zero-shot ASR outputs, we investigate whether they are effective at correcting fine-tuned ASR outputs. To this end, we generate fine-tuned ASR outputs for both training and test sets. We conduct 2-fold cross-validation on the training set and use the validation ASR outputs as the training set for LLM error correction. For the test sets, we use the fine-tuned ASR models trained on the full training set. We use WSP-L-T and WavLM-L as the ASR models. 

\section{Experiments}
\subsection{Evaluation}
We report mean Word Error Rate (WER) across all utterances.
Before calculating WER for each utterance, we pass the ground truth transcript and ASR outputs to the Whisper normalizer.

\subsection{Dataset}
We consider two child conversational datasets: My Science Tutor (MyST) Children’s speech corpus~\cite{pradhan2024my} and ADOS-Mod3 corpus of Autism diagnostic administration \cite{lahiri2022interpersonal}. Our research complies with all Institutional Review Board (IRB) protocols and follows the Data Use Agreements (DUAs) established by the original data providers.

\noindent \textbf{MyST} \cite{ward2011my} includes transcribed conversations between children and virtual tutors. The children were recruited from grade 3 to grade 5, which corresponds to around 8 to 12 years of age. The topics include 8 areas of science, such as biology, physics, and others. Similar to \cite{fan24b_interspeech}, we filter out samples longer than 30s (maximum length for Whisper). However, unlike benchmark results reported from \cite{fan24b_interspeech}, we did not filter out the samples based on the number of words or WER from zero-shot Whisper-large, allowing us to get a more comprehensive assessment of the ASR performance. In addition, we use this dataset only for the experiments for ASR Error Correction \textit{without} Context, as the corpus does not include speech or transcript data from the virtual tutors. We use the official training and test split for the LLM ASR error correction experiments.

\noindent \textbf{ADOS-Mod3} dataset \cite{lahiri2022interpersonal} contains 352 sessions collected from 180 children during two specific sections of the ADOS-2 autism diagnostic protocol: ``Social Difficulties'' and ``Annoyance and Emotional'' tasks. The children ranged in age from 2 to 13 years, with 45 being female. Approximately half received an autism spectrum disorder diagnosis, while the remaining children were diagnosed with various other conditions, including ADHD and mental or language disorders. 96 children and 84 children were recorded at two different medical centers in Chicago (CHIC) and Michigan (MICH). On average, each session contains 25.9 child utterances and 30.0 adult utterances, with mean durations of 2.58s and 2.06s, respectively. For LLM-based ASR error correction experiments, we use data collected from CHIC as the training set and MICH as the test set. Since this dataset contains both child and adult speech utterances, we also report the individual WERs.

\subsection{Experimental Setup: ASR Fine-tuning}
For Whisper finetuning, we choose WSP-L-T because of its performance and relatively smaller size compared to WSP-L. We train for 2000 steps with a learning rate of $1e-6$. For the WavLM, we train for 30000 steps with a learning rate of $3e-4$. Adam optimizer is used with a batch size of 32 for both models. The same configurations are used for both datasets. We choose the best model based on the validation WER.

For the MyST dataset, we use the official training and validation sets for fine-tuning, and we report the WERs on the test set. However, to prepare training data for the LLM instruction tuning for fine-tuned ASR models, we split the training data in half for fine-tuning ASR models. We then apply the fine-tuned ASR models to the other half of the training dataset. We conduct the fine-tuning for each of the 2-splits.

For the ADOS-MOD3 dataset, we randomly select $80\%$ from the CINC as the training set and the rest of $20\%$ as the validation set, with the data from MICH as the test set. Similar to the MyST dataset setup, we conduct a 2-split fine-tuning to generate the fine-tuned ASR outputs for the train set.

\subsection{Experimental Setup: LLM Instruction tuning}

We experiment with the LLaMa 3.1-8B and LLaMa 3.2-1B models for ASR correction. We train the LLMs for 5 and 10 epochs on the MyST and the ADOS datasets, respectively. We apply a learning rate of $5e-4$ in all LLM fine-tuning experiments. Our system prompt for fine-tuning the ADOS dataset is shown in Figure~\ref{fig:no_context}, while the system prompt for fine-tuning the MyST dataset is ``You are a helpful assistant that helps to correct transcriptions from a child in a tutoring session." We empirically tested that ASR correction results remain robust across varying temperature values, with different temperatures yielding similar ASR correction outputs. Therefore, we set the temperature to 0.2 during the inference phase in all experiments. Even though LLMs generally produce reasonable outputs, we identify instances where they could generate repeated or hallucinated lengthy content. Thus, we set the ASR output as the best hypothesis whenever the generated output exceeds the best ASR hypothesis by more than three words.

\section{Results and Discussion}

  
 
\begin{table}[t]
  \centering
  \caption{WER comparison with LLM for zero-shot ASR error correction using ADOS-Mod3 and MyST dataset.}
  \vspace{-3mm}
  \footnotesize
  \begin{tabular}{l c c c c c}
    \toprule
    \multirow{2}{*}{\textbf{ASR}} & 
    \multirow{2}{*}{\textbf{LLaMA3}} &
    \multicolumn{3}{c}{\textbf{ADOS}} & 
    \multicolumn{1}{c}{\textbf{MyST}} \\ 

    & &
    \textbf{Overall} & 
    \textbf{Child} &
    \textbf{Adult} & 
    \textbf{Child} \\ 
    \cmidrule(lr){1-1} \cmidrule(lr){2-2} \cmidrule(lr){3-5} \cmidrule(lr){6-6}
    \multirow{2}{*}{WSP-S} & Unused & $46.67$ & $63.73$ & $32.23$ & $22.33$ \\
    & 1B & $47.19$ & $64.64$ & $32.41$ & $22.20$  \\
     & 8B & $\mathbf{43.96}$ & $\mathbf{62.71}$ & $\mathbf{28.10}$ & $\mathbf{20.60}$ \\
     \cmidrule(lr){1-1} \cmidrule(lr){2-2} \cmidrule(lr){3-5} \cmidrule(lr){6-6}
     \multirow{2}{*}{WSP-L-T} & Unused & $40.77$ & $55.84$ & $28.07$ & $20.01$ \\
     & 1B & $39.11$ & $54.29$ & $26.30$ & $19.66$ \\
     & 8B & $\mathbf{37.09}$ & $\mathbf{53.87}$ & $\mathbf{22.94}$  & $\mathbf{18.35}$ \\
     \cmidrule(lr){1-1} \cmidrule(lr){2-2} \cmidrule(lr){3-5} \cmidrule(lr){6-6}
     \multirow{2}{*}{WSP-L} & Unused & $40.26$ & $55.19$ & $27.65$ & $19.58$ \\
     & 1B & $39.55$ & $54.48$ & $26.93$ & $19.50$ \\
     & 8B & $\mathbf{36.70}$ & $\mathbf{52.63}$ & $\mathbf{23.24}$ & $\mathbf{18.41}$\\
    \bottomrule
  \end{tabular}
  
  \label{tab:zero_ados}
\end{table}

\begin{table}[t]
  
  \centering
  \caption{WER comparison with LLM for fine-tuned ASR output error correction using ADOS-Mod3 and MyST dataset.}
  \vspace{-3mm}
  \footnotesize
  \begin{tabular}{l c c c c c}
    \toprule
    \multirow{2}{*}{\textbf{ASR}} & 
    \multirow{2}{*}{\textbf{LLaMA3}} &
    \multicolumn{3}{c}{\textbf{ADOS}} & 
    \multicolumn{1}{c}{\textbf{MyST}} \\ 

    & &
    \textbf{Overall} & 
    \textbf{Child} &
    \textbf{Adult} & 
    \textbf{Child} \\ 
    \cmidrule(lr){1-1} \cmidrule(lr){2-2} \cmidrule(lr){3-5} \cmidrule(lr){6-6}
    \multirow{2}{*}{WSP-L-T} & Unused & $\mathbf{32.11}$ & $\mathbf{46.99}$ & $\mathbf{19.47}$ & $\mathbf{14.31}$\\
     & 1B & $33.25$ & $47.93$ & $20.77$ & $14.55$\\
     & 8B & $32.92$ & $47.47$ & $20.56$ & $14.40$\\
     
    \cmidrule(lr){1-1} \cmidrule(lr){2-2} \cmidrule(lr){3-5} \cmidrule(lr){6-6}
    \multirow{2}{*}{WavLM-L} & Unused & $66.33$ & $88.05$ & $47.87$ & $27.54$ \\
     & 1B & $56.83$ & $78.34$ & $38.54$ & $19.93$ \\
     & 8B & $\mathbf{50.58}$ & $\mathbf{72.24}$ & $\mathbf{16.45}$ & $\mathbf{16.45}$\\
     
    \bottomrule
  \end{tabular}
  
  \label{tab:finetuned_ados}
\end{table}

\subsection{Can LLMs Improve \textit{zero-shot} Child ASR Results?}
\label{sec:zero_results}
Table~\ref{tab:zero_ados} shows the LLM error correction results for the zero-shot Whisper ASR outputs. We see consistent reductions in WERs across all three ASR models for each dataset when LLaMA 3.1-8B model is used. The improvements are less substantial when the LLaMA 3.2-1B model is applied. Interestingly, the 3.2-1B model slightly increases the WER for WSP-S with the ADOS-Mod3 dataset. Thus, the parameter size of the LLM used is critical for the ASR error correction in this domain, and larger LLMs tend to perform better in ASR corrections. In summary, these results show that when the audio resources or the ASR models are unavailable for training, LLMs with larger parameter sizes can help refine the transcriptions.

\subsection{Can LLMs Improve \textit{Fine-tuned} Child ASR Results?}
Table~\ref{tab:finetuned_ados} shows WERs from the fine-tuned ASR outputs before and after applying LLM error corrections. Since the previous benchmark applied additional dataset filtering (e.g, removing utterances with WER $> 50\%$), the WER for MyST in this study is higher than those reported in \cite{fan24b_interspeech}. The results show that the WSP-L-T model substantially outperforms WavLM-L for both datasets, similar to \cite{fan24b_interspeech}. Based on our manual error inspections, one plausible reason is that the CTC-based character prediction approach produces more spelling errors, especially for children with underdeveloped pronunciation capabilities. In contrast, the Whisper model is less likely to make spelling mistakes since it generates the transcriptions by predicting byte-level Byte-Pair Encoding (BPE) tokens autoregressively during inference.  

For both datasets, LLMs show substantial improvements for WavLM ASR outputs, while they do not show improvements for Whisper ASR outputs. We observe that LLMs help correct spelling errors that the fine-tuned WavLM produces. However, we reason that LLMs show limited advantages for Whisper outputs because both systems use similar autoregressive decoding, where each token depends on previous predictions. In addition, unlike the Whisper decoder, LLM decoders can not access the speech features through cross-attention.

\subsection{Does Context Improve LLM Error Correction?}
Table~\ref{tab:context} presents the WERs obtained using LLM-based ASR error correction with contextual information from the previous 1 or 3 utterances. Contrary to our initial hypothesis, incorporating previously predicted utterances leads to increased WERs compared to LLM-based ASR error correction without context, across all our experimental conditions. Furthermore, using the context of 3 utterances yields higher error rates than using the context of a single utterance. \textbf{One plausible reason behind this performance degradation is error propagation}, as the previously predicted utterances already include recognition errors. 

\begin{table}[t]
  
  \centering
  \caption{WER comparison with LLaMA 3.1-8B for ASR error correction using context. ADOS-Mod3 dataset is used. (ft) indicates whether the ASR model is fine-tuned or not.}
  \vspace{-3mm}
  \footnotesize
  \begin{tabular}{l c c c c}
    \toprule
    \textbf{ASR (ft)} & 
    \textbf{\# Context} & 
    \textbf{Overall} & 
    \textbf{Child} &
    \textbf{Adult} \\ 
    \cmidrule(lr){1-1} \cmidrule(lr){2-2} \cmidrule(lr){3-5}
     WSP-L-T (No) & 1 & $38.06$ & $55.67$ & $23.21$ \\
     & 3 & $37.79$ & $54.65$ & $23.56$ \\
    \cmidrule(lr){1-1} \cmidrule(lr){2-2} \cmidrule(lr){3-5}
     WSP-L-T (Yes) & 1 & $33.02$ & $47.03$ & $21.1$ \\
     & 3 & $37.87$ & $55.58$ & $22.81$ \\
     \cmidrule(lr){1-1} \cmidrule(lr){2-2} \cmidrule(lr){3-5}
     WavLM-L (Yes) & 1 & $52.99$ & $74.46$ & $34.73$ \\
     & 3 & $54.98$ & $78.47$ & $35.02$ \\
    \bottomrule
  \vspace{-6mm}
  \end{tabular} 
  \label{tab:context}
\end{table}

\begin{figure}[!t]
  \centering
  \centerline{\includegraphics[width=0.85\linewidth]{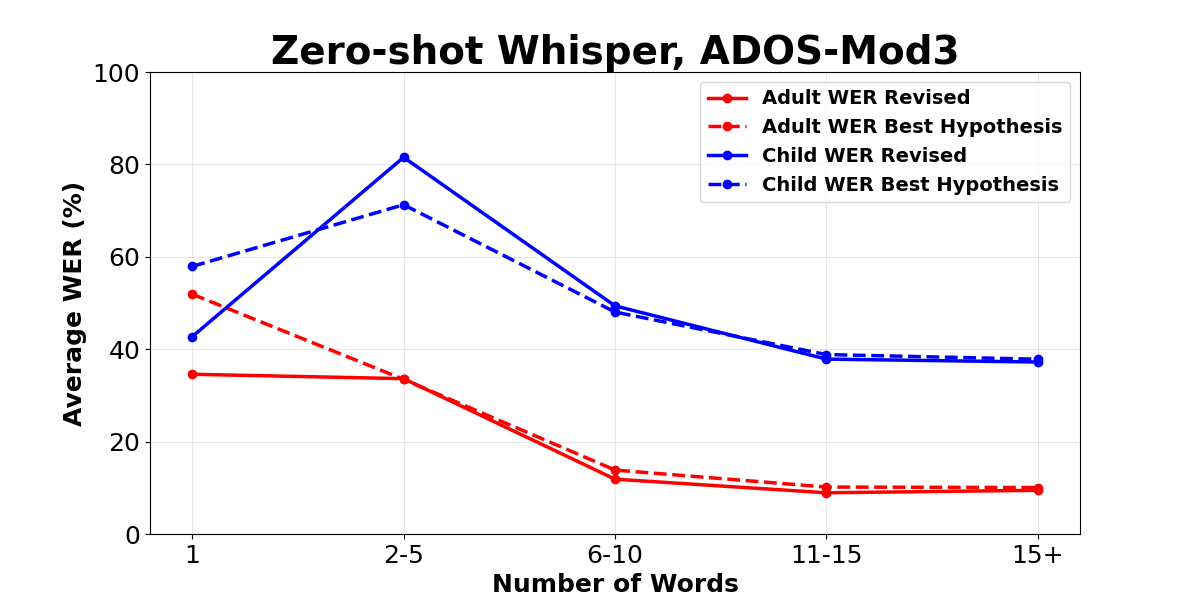}}
  \centering
  \centerline{\includegraphics[width=0.85\linewidth]{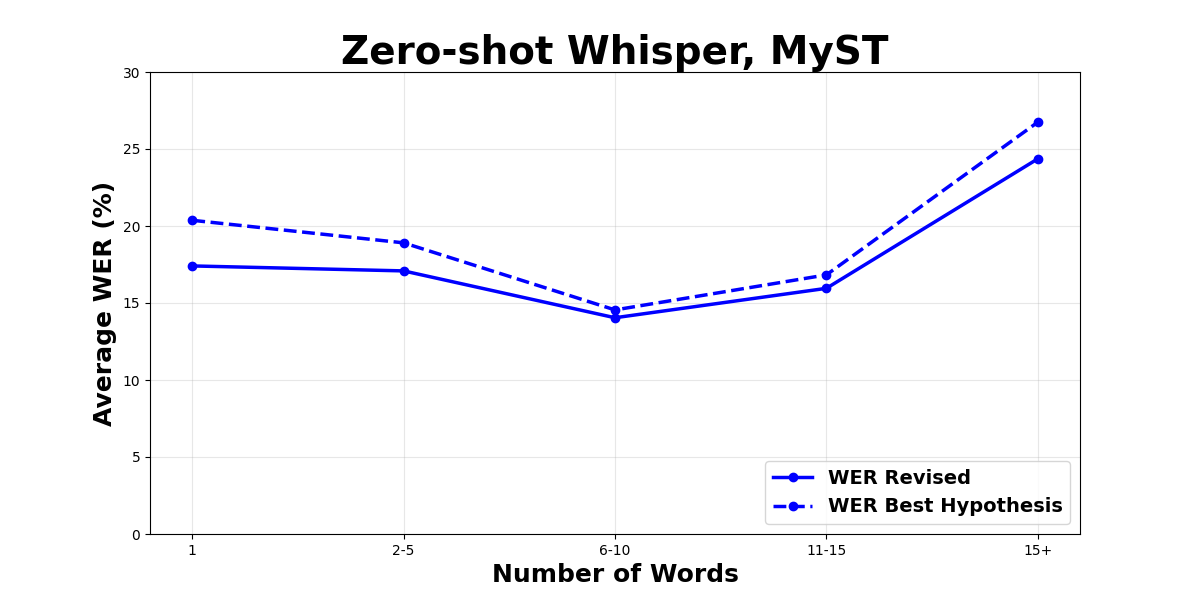}}
  \caption{WERs by utterance lengths with zero-shot Whisper ASR (WSP-L-T). Results from both datasets.}\medskip
  \label{fig:length_zero}
  \vspace{-5mm}
\end{figure}

\subsection{Analysis on utterance length}
To have a more comprehensive understanding of when LLMs can help correct ASR errors, we examine how the WER changes when varying the number of words in the utterances. Figure \ref{fig:length_zero} shows the results for zero-shot ASR with WSP-L-T using both ADOS-Mod3 and MyST datasets. The zero-shot ASR results show that LLM correction is most effective for single-word utterances. This is likely because Whisper models often generate phonetically similar but contextually inappropriate words (including words from other languages) when processing short or unclear utterances. The language understanding of the LLMs helps filter out these mismatched transcriptions and convert them to more probable conversational utterances. We have also found that the improvements are more observable for adult speech in ADOS-Mod3 and child speech in MyST than for child speech in ADOS-Mod3. This is likely due to the challenges in correcting ASR outputs of children with less developed language skills prevalent in ADOS-Mod3 data.

\begin{figure}[!t]
  \centering
  \centerline{\includegraphics[width=0.85\linewidth]{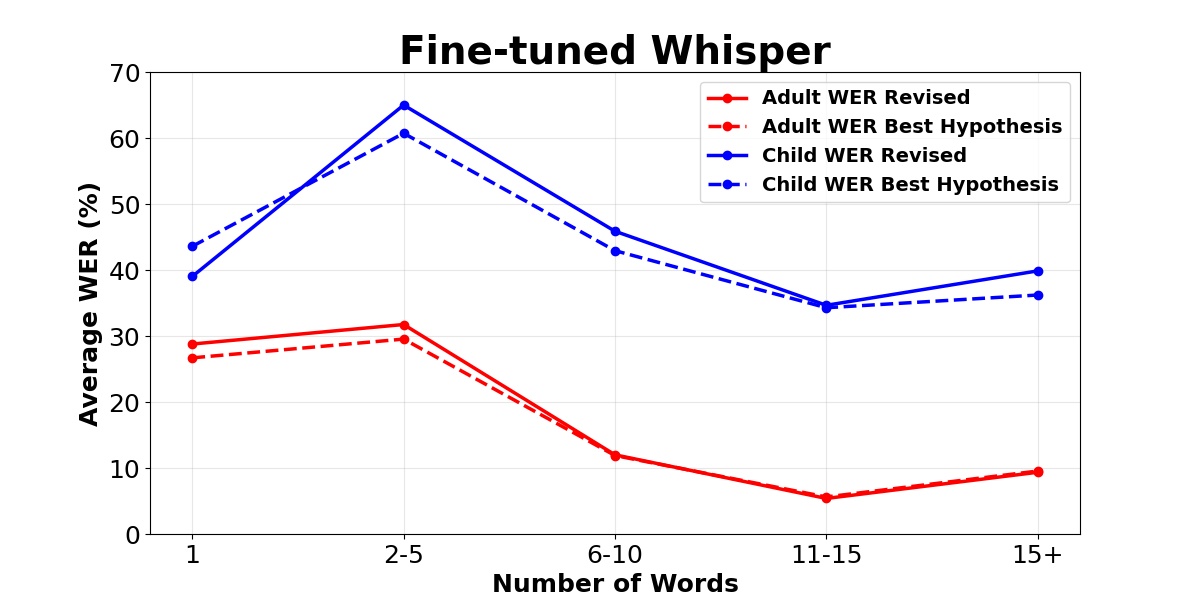}}  \centering
  \centerline{\includegraphics[width=0.85\linewidth]{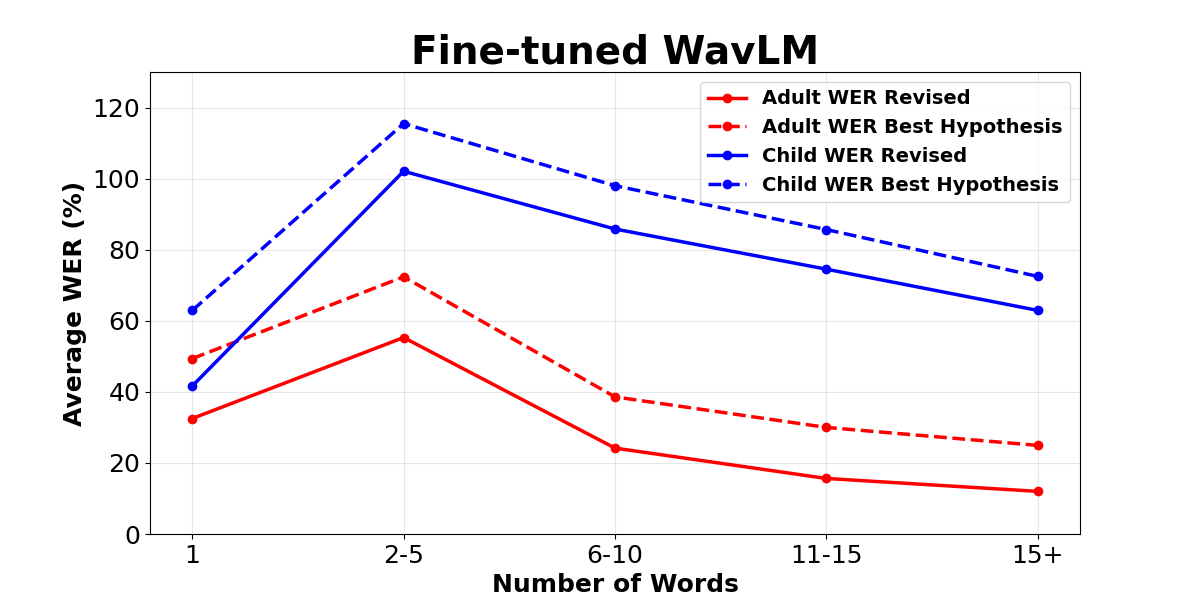}}
  \caption{WERs by utterance lengths with fine-tuned ASR models (WSP-L-T, WavLM-L), using the ADOS-Mod3 dataset.}\medskip
  \label{fig:length_ft}
  \vspace{-7mm}
\end{figure}

Moreover, Figure~\ref{fig:length_ft} shows the results for fine-tuned ASR with WSP-L-T and WavLM-L, using only the ADOS-Mod3 dataset. The results demonstrate that LLMs do not show improvements for the Whisper model other than for single utterances from children. For WavLM, the improvements are consistent across utterances with different utterance lengths, where we found most of the improvements are from correcting spelling errors as discussed in Section \ref{sec:zero_results}.

\section{Conclusion}
This paper has investigated the use of LLMs for ASR error correction in child conversations, making several key findings. First, larger LLMs consistently improve zero-shot ASR performance across different Whisper models, while smaller LLMs show limited benefits. Second, for fine-tuned ASR systems, LLMs substantially improve CTC-based self-supervised ASR outputs by correcting spelling errors but show minimal improvements for the outputs from Whisper, a supervised ASR model with the attention-based encoder-decoder architecture. Third, contrary to initial hypotheses, incorporating conversational context degrades error correction performance, likely due to error propagation from previous utterances. These findings advance our understanding of LLM capabilities in child speech recognition while highlighting some of its limitations. Future work may include investigating larger LLMs and designing more effective strategies to incorporate conversational context.

\section{Acknowledgment}

This work was supported by \textsc{Simons Foundation (SFI-AR-HUMAN-00004115-03, 655054)}.

\bibliographystyle{IEEEtran}
\bibliography{mybib}

\end{document}